\documentclass[letterpaper]{article} 
\usepackage{aaai2026}  
\usepackage{amssymb}
\usepackage{amsmath}
\usepackage{multirow}
\usepackage{pifont}
\newcommand{\cmark}{\ding{51}}  
\newcommand{\xmark}{\ding{55}}  
\usepackage[table,xcdraw]{xcolor}
\usepackage{booktabs}
\usepackage{times}  
\usepackage{helvet}  
\usepackage{courier}  
\usepackage[hyphens]{url}  
\usepackage{graphicx} 
\usepackage{natbib}  
\usepackage{caption} 
\usepackage{algorithm}
\usepackage{algorithmic}

\usepackage{newfloat}
\usepackage{listings}
\DeclareCaptionStyle{ruled}{labelfont=normalfont,labelsep=colon,strut=off} 
\floatstyle{ruled}
\newfloat{listing}{tb}{lst}{}
\floatname{listing}{Listing}
\title{DeformTrace: A Deformable State Space Model\\with Relay Tokens for Temporal Forgery Localization}
\author {
    Xiaodong Zhu,
    Suting Wang,
    Yuanming Zheng,
    Junqi Yang,
    Yangxu Liao,\\
    Yuhong Yang\thanks{Corresponding author.},
    Weiping Tu,
    Zhongyuan Wang
}
\affiliations {
    National Engineering Research Center for Multimedia Software, School of Computer Science, Wuhan University, China\\
    xiaodongzhu@whu.edu.cn, yangyuhong@whu.edu.cn, wzy\_hope@163.com
}

\usepackage{bibentry}

\begin{document}

\maketitle

\begin{abstract}
Temporal Forgery Localization (TFL) aims to precisely identify manipulated segments in video and audio, offering strong interpretability for security and forensics. While recent State Space Models (SSMs) show promise in precise temporal reasoning, their use in TFL is hindered by ambiguous boundaries, sparse forgeries, and limited long-range modeling. We propose DeformTrace, which enhances SSMs with deformable dynamics and relay mechanisms to address these challenges. Specifically, Deformable Self-SSM (DS-SSM) introduces dynamic receptive fields into SSMs for precise temporal localization. To further enhance its capacity for temporal reasoning and mitigate long-range decay, a Relay Token Mechanism is integrated into DS-SSM. Besides, Deformable Cross-SSM (DC-SSM) partitions the global state space into query-specific subspaces, reducing non-forgery information accumulation and boosting sensitivity to sparse forgeries. These components are integrated into a hybrid architecture that combines the global modeling of Transformers with the efficiency of SSMs. Extensive experiments show that DeformTrace achieves state-of-the-art performance with fewer parameters, faster inference, and stronger robustness.
\end{abstract}


\section{Introduction}

The rapid development of generative AI has greatly advanced multimedia creation but also raised security concerns due to the ease of producing realistic forgeries. Most existing work \cite{feng2023self, oorloff2024avff, smeu2025circumventing} focuses on binary forgery detection, whereas Temporal Forgery Localization (TFL) provides finer interpretability by identifying manipulated segments.

\begin{figure}[t]
\centering
\includegraphics[width=\columnwidth]{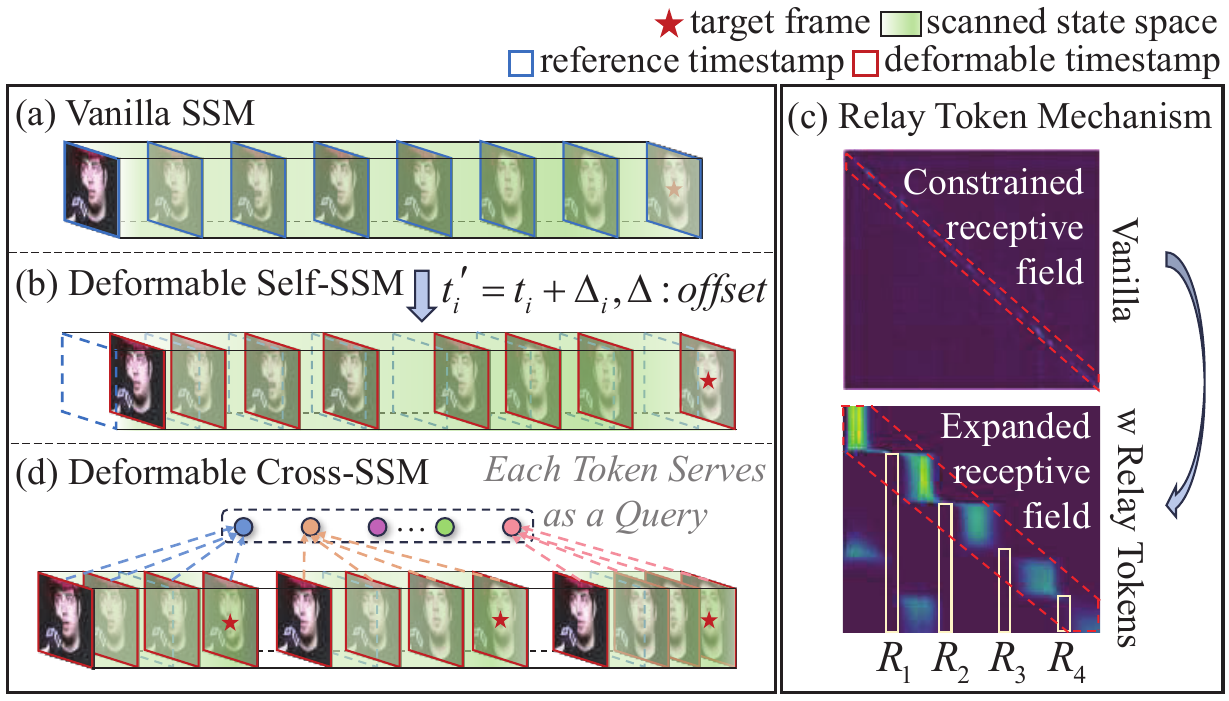} 
\caption{Overview of our main contributions (we take the video sequence for illustration). (a) Vanilla SSM; (b) Deformable Self-SSM with learnable temporal offsets for flexible local sampling; (c) Visualization of hidden attention, where relay tokens expand receptive fields and maintain long-range dependencies; (d) Deformable Cross-SSM enables cross-sequence interactions by allowing each query token to partition the global state space into subspaces.}
\label{teaser}
\end{figure}

Early TFL methods, such as BA-TFD \cite{cai2022you} and BA-TFD+ \cite{cai2023glitch}, rely on CNNs or multi-scale transformers for frame-level classification, but suffer from low precision and slow inference. Later models like UMMAFormer \cite{zhang2023ummaformer} and DiMoDif \cite{koutlis2024dimodif} improve performance with pretrained extractors and feature pyramids, yet remain computationally heavy. Thus, developing a compact yet effective TFL architecture is imperative.TFL is conceptually related to temporal action detection (TAD), as both require precise boundary localization in long sequences. Recent progress in TAD using state space models (SSMs) and query-based designs has provided valuable insights for model design.

Recent advances in State Space Models (SSMs), especially Mamba \cite{mamba, mamba2}, demonstrate strong performance with lower complexity and faster inference. Models such as MambaIRv2 \cite{guo2025mambairv2} and ActionMamba \cite{chen2024video} outperform baselines in tasks like segmentation and temporal action detection, showing the potential of SSMs for dense, temporally-aware prediction. However, SSM-based architectures have not been applied to TFL due to three main challenges: \textbf{Boundary ambiguity}: Forgery boundaries are often unclear, unlike the well-defined ones in TAD. Standard SSMs use fixed state updates, causing temporal smoothing that reduces localization precision; \textbf{Sparse forgeries}: Most frames are non-forged, so SSMs’ recursive updates are dominated by non-forgery patterns, weakening sensitivity to sparse forgeries; \textbf{Limited long-sequence modeling}: Although efficient for long sequences, SSMs suffer from information decay over distance \cite{ye2025longmamba}, limiting their ability to capture long-range context.

To address boundary ambiguity, we propose Deformable Self-SSM (DS-SSM), which, for the first time, integrates a deformable dynamic receptive field mechanism into state space models. Unlike deformable Mamba variants in the image domain \cite{liu2025defmamba, hu2024deformable}, our design leverages inherent temporal continuity of video and audio, thus omits operations such as patch splitting and token ranking. Instead, DS-SSM predicts offsets at each time step to dynamically sample input features, significantly reducing computational overhead while preserving SSMs’ low complexity. Figure~\ref{teaser}(b) shows that, compared to vanilla SSM (see Figure~\ref{teaser}(a)), DS-SSM better captures semantically relevant context beyond local fixed windows and improves robustness to ambiguous temporal boundaries.

Meanwhile, we introduce a Relay Token Mechanism into the DS-SSM to maintain long-range information flow, inspired by relay nodes in wireless communication. As shown in Figure~\ref{teaser}(c), which compares the hidden attention of vanilla SSM and our variant with relay tokens (visualized following \cite{ali2025hidden}), these periodically inserted learnable tokens effectively expands the receptive field and mitigates the long-range decay problem.

We further propose the Deformable Cross-SSM (DC-SSM), which introduces cross-sequence interactions into deformable state space modeling to tackle sparse forgeries. As shown in Figure~\ref{teaser}(d), each auxiliary token, representing a potential forgery, serves as a query to retrieve forgery-relevant information from the main stream sequence. This mechanism partitions the global state space into query-specific subspaces, reducing non-forgery accumulation and improving sensitivity to sparse forgeries. Like cross-attention, DC-SSM enables explicit token-to-sequence interactions for targeted retrieval. Finally, These components are integrated into a hybrid TFL architecture combining Transformer attention with efficient state updates.

To sum up, the main contributions of this work are: (1) We propose Deformable Self-SSM (DS-SSM), the first to introduce dynamic receptive fields into temporal state space models, improving localization of ambiguous boundaries. (2) We introduce a Relay Token mechanism that explicitly mitigates the long-range decay of SSMs, which is a key limitation in prior state-space models. (3) We propose Deformable Cross-SSM (DC-SSM), the first to incorporate cross-sequence interactions into state space modeling, enhancing sensitivity to sparse forgeries.  (4) We integrate these components into a unified hybrid TFL architecture that combines the strengths of Transformers and SSMs. Extensive experiments on standard benchmarks demonstrate state-of-the-art performance with lower model size and faster inference.

\begin{figure*}[t]
\centering
\includegraphics[width=0.88\textwidth]{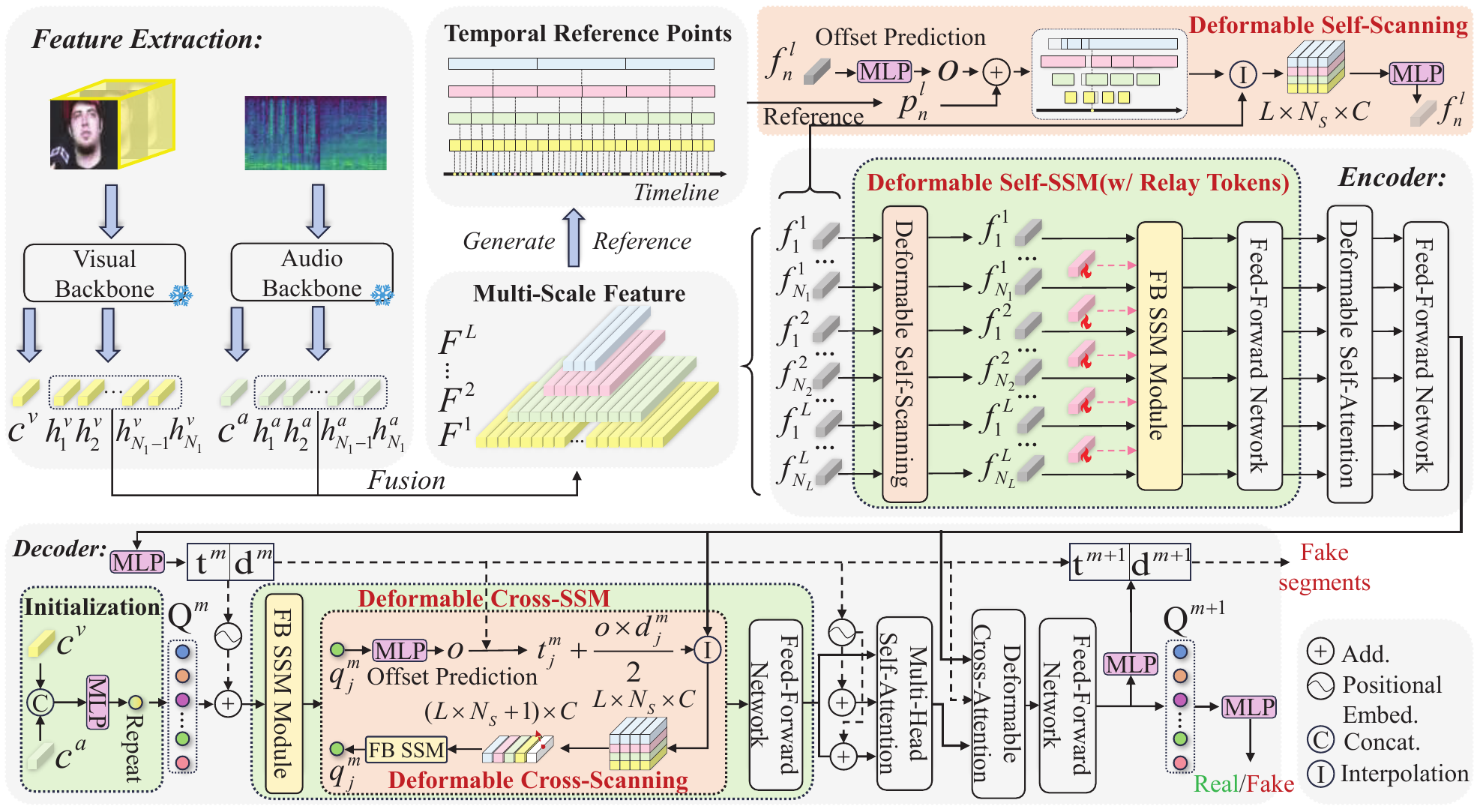} 
\caption{Illustration of the overall scheme of DeformTrace. Built on TadTR \cite{liu2022end}, DeformTrace integrates a multi-scale audio-visual feature extraction module, a deformable encoder for temporal modeling, and a deformable decoder for forgery localization and video-level classification. By incorporating deformable self- and cross-SSM modules, it combines Mamba’s efficient state updates with Transformer's global modeling. Relay tokens with enhanced and cooperation losses help preserve long-range information dependencies during self-scanning.}
\label{overall}
\end{figure*}

\section{Related Work}

\subsection{State Space Model Applications}

State Space Models (SSMs) provide an efficient approach to sequence modeling. Mamba~\cite{mamba, mamba2} builds on this by introducing a selective mechanism that enables long-sequence modeling with linear complexity. Mamba has inspired a growing number of variants in vision tasks, focusing on scalability, efficiency, and local-global modeling \cite{vim, liu2024vmamba, pei2025efficientvmamba, patro2024simba, hatamizadeh2025mambavision}. Meanwhile, deformable variants \cite{liu2025defmamba, hu2024deformable} incorporate deformable scanning to handle irregular patterns but they are restricted to uni-sequence modeling. Moreover, both Mamba and its variants suffer from long-range decay. LongMamba \cite{ye2025longmamba} attempts to solve this using channel classification and token filtering, but its reliance on fixed thresholds limits adaptability. 

\subsection{Temporal Action Detection}

Temporal Action Detection (TAD) aims to detect and localize actions in untrimmed videos. Early anchor-based methods \cite{bai2020boundary, lin2018bsn, lin2019bmn, long2019gaussian} struggle with varying action lengths due to fixed anchors. Anchor-free methods \cite{cheng2022tallformer, lin2021learning, zhang2022actionformer} improve flexibility via center-based or asymmetric designs. Query-based approaches like TadTR \cite{liu2022end} and TE-TAD \cite{kim2024te}, inspired by DETR \cite{carion2020end}, treat TAL as set prediction, removing anchors and post-processing \cite{tan2021relaxed}. Recently, Mamba-based architectures \cite{chen2024video, liu2024harnessing, wang2025temporal} have been introduced to TAD, enhancing temporal modeling. 

\subsection{Temporal Forgery Localization}

Temporal Forgery Localization (TFL) aims to identify forged segments in videos, requiring fine-grained temporal modeling. While most research focuses on video-level detection \cite{zhou2021joint, yang2023avoid, yu2023pvass}, TFL remains underexplored. Recent datasets such as Lav-DF \cite{cai2023glitch} and AV-Deepfake1M \cite{cai2024av} have started to drive progress in this direction. BA-TFD \cite{cai2022you} introduced contrastive loss for frame-level classification, followed by BA-TFD+ \cite{cai2023glitch} with multi-scale Transformers. UMMAFormer \cite{zhang2023ummaformer}, DiMoDif \cite{koutlis2024dimodif}, MFMS \cite{zhang2024mfms}, and CLFormer \cite{cheng2025clformer} improved accuracy via attention and feature pyramids. Vigo \cite{perez2024vigo} employs an IoU-based fusion strategy, while TransHFC \cite{huang2025transhfc} integrates hypergraph learning and Transformers. Zhu et al. \cite{zhu2025query}proposed a query-based audio-visual framework for efficient end-to-end temporal localization.

\section{The Proposed Method}

\subsection{Overall Architecture}

Given a dataset of untrimmed talking face videos, each sample \(x = \{x^v, x^a\}\) contains temporally aligned video and audio streams. However, parts of \(x\) may be manipulated in either modality. The goal of DeformTrace is to localize forgery segments \(\mathcal{A} = \{(t_i, d_i)\}_{i=1}^{N_f}\), where \(t_i\) and \(d_i\) represent the center timestamp and duration of the \(i\)-th forgery, respectively. Here, \(N_f\) denotes the total number of forgeries. Additionally, it predicts a binary authenticity label \(y\).

\textbf{Feature Extraction.} As illustrated in Figure~\ref{overall}, DeformTrace adopts a query-based architecture based on TadTR~\cite{liu2022end}. The video \(x^v \in \mathbb{R}^{T \times H \times W \times C_0}\) and the mel spectrogram \(x^a \in \mathbb{R}^{T \times N_{\text{mel}}}\) are processed by frozen, pretrained visual and audio backbones to extract classification tokens \(\{c^v, c^a\}\) and feature sequences \(\{h_n^v, h_n^a\}_{n=1}^{N_1}\), where both the classification and feature tokens have a dimensionality of \(C\). Here, \(T\) and \(N_1\) represent the number of input frames and extracted tokens of the first scale. To preserve temporal resolution for fine-grained localization, we set \(T=N_1\). These sequences are concatenated and linearly projected to form fused features \(F^1\), further processed by \((L-1)\) \(2\times\)downsampling layers to obtain features \(\mathcal{F}=\{F^l\}_{l=1}^L\), capturing multi-scale temporal context.

\textbf{Encoder.} The encoder receives flattened features \(\{f_n^l\}_{n=1}^{N_l}\), where \(n\) denotes the feature index of each scale and \(N_l = N_1 / 2^{l-1}, l=1,2,\cdots,L\), then the encoder outputs a context-enhanced sequence \(X_E\). As shown in Figure~\ref{overall}, it comprises a Deformable Self-SSM, a Deformable Self-Attention (DSA)~\cite{liu2022end}, and a Feed-Forward Network (FFN). The DSA adaptively attends to sparse temporal locations around a reference point, while the FFN adds nonlinearity and further refines features.

\textbf{Decoder.} Visual and audio classification tokens are fused via an MLP into a global multimodal feature, which is repeated \(N_q\) times to form initial queries \(Q^0 = \{q_j^0\}_{j=1}^{N_q}\), providing coarse alignment with ground-truth and aiding convergence~\cite{kim2024te}. Initial proposals \(\mathcal{A}^0 = \{(t^0_j, d^0_j)\}_{j=1}^{N_q}\), representing center and duration of segments, are generated from fused features via MLP. The decoder has \(M\) layers, each containing Deformable Cross-SSM, Multi-Head Self-Attention (MHSA), Deformable Cross-Attention (DCA)~\cite{liu2022end}, and FFN. MHSA models inter-query relations using anchor-based positional embeddings, while DCA allows each query to attend to relevant audio-visual context. The DCA output is passed through the FFN to update queries and then through an MLP to predict offsets \(\Delta t, \Delta d\) for proposal refinement. The final proposals \(\mathcal{A}^M\) are used as predicted forgery segments, and the final queries \(Q^M\) are passed through an MLP classifier to produce the video-level forgery score \(\hat{y}\).

\subsection{Deformable Self-SSM} 

The Deformable Self-SSM (DS-SSM) module comprises three components: Deformable Self-Scanning, Forward-Backward SSM (FB SSM) \cite{wang2022pretraining}, and a Feed-Forward Network (FFN). Deformable Self-Scanning enables each token to flexibly sample semantically relevant positions beyond local windows. FB SSM performs bidirectional state-space propagation on the sampled sequence, capturing contextual information from both past and future.

For the \(n\)-th feature in the \(l\)-th layer of the multi-scale feature \(\mathcal{F}\), denoted \(f_n^l\), we compute a normalized reference location \(p_n^l\) to align with the temporal axis of the video:
\begin{equation}
p_n^l = \frac{\omega \cdot 2^{l-1} \cdot (n + 0.5)}{\text{fps} \cdot d}.
\label{eq:1}
\end{equation}
where \(\text{fps}\) denotes frames per second (default 25), \(\omega\) is the stride that indicates how many frames to step when extracting features, and \(d\) is the video duration in seconds. The numerator above estimates the temporal center of \(f_n^l\) in terms of frame index, converts it to seconds by dividing by \(\text{fps}\), and normalizes the result by \(d\), yielding a value in \((0, 1]\). We pre-compute these temporal reference points to accelerate the sampling process.

Given a feature \(f_n^l\) and its temporal reference point \(p_n^l\), we predict an offset matrix \(o \in \mathbb{R}^{L \times N_s}\) via a multi-layer perceptron (MLP), where \(N_s\) is the number of offsets per scale. For each feature, multiple offsets are predicted across scales, enhancing the flexibility and diversity of sampling. \(p_n^l\) is repeated and added element-wise with \(o\) to obtain deformable points \(\hat{p}\). A bilinear interpolation function \(\phi(\cdot,\cdot)\) extracts features at positions \(\hat{p}\) from multi-scale feature \(\{F^l\}_{l=1}^L\), which are then aggregated via another MLP to produce the output feature \(\hat{f}_n^l\). The process is formulated as:

\begin{equation}
\begin{aligned}
o &= \operatorname{MLP}_1(f_n^l), \\
\hat{p} &= \operatorname{Repeat}(p_n^l) + o, \\
\hat{f}_n^l &= \operatorname{MLP}_2\left( \phi\left(\hat{p}, \{F^l\}_{l=1}^L \right) \right).
\end{aligned}
\label{eq:2}
\end{equation}

\subsection{Relay Token Mechanism}

Long-range decay is an inherent limitation of SSMs. Token interactions rely on the control matrix power \(\bar{A}^k\), where \(k\) is the pairwise distance \cite{mamba, mamba2}. Since \(\bar{A}\) contains values less than 1, interactions decay exponentially with distance. For instance, in DS-SSM, semantically aligned tokens \(f_1^1\) and \(f_1^L\) are separated by \(k \approx 2N_1\) (with \(N_1 > 100\) in datasets like AV-Deepfake1M and LAV-DF), resulting in negligible interaction and limited receptive fields (see Figure~\ref{teaser}(d)).

To address this, we draw inspiration from communication systems, where relay nodes mitigate long-distance signal attenuation. We propose a Relay Token Mechanism: learnable global tokens \(\{r_i\}_{i=1}^{N_r}\) are evenly inserted into the input sequence before state-space updates. These input-independent tokens partition the sequence into \(N_r + 1\) subspaces. Within each subspace, local states efficiently pass information to the relay token, which then broadcasts the aggregated message to other subspaces. This forms a sparse sequence-to-token information flow across the sequence. Figure~\ref{teaser}(d) shows attention maps illustrating receptive field expansion and information relay via the relay tokens.

\subsection{Deformable Cross-SSM}

The Deformable Cross-SSM (DC-SSM) module comprises three components. The FB-SSM captures temporal dependencies around each query, while Deformable Cross-Scanning allows each query to dynamically attend to semantically relevant audio-visual features from the encoder, thereby enhancing the localization of forged regions. An FFN further refines the resulting features.

In Deformable Cross-Scanning, for the \(m\)-th layer of the decoder, each query \(q_j^m\in Q^m\) uses its anchor proposal \((t_j^m, d_j^m)\) as a reference. A MLP predicts a multi-scale offset \(o \in \mathbb{R}^{L \times N_s}\), and the deformable points are computed as \(p_j^m = t_j^m + o \times d_j^m / 2\). A bilinear interpolation function samples features at \(p_j^m\) from the encoder output \(X_E\), producing a feature tensor of shape \(L \times N_s \times C\). This tensor is flattened and concatenated with a learnable empty token, forming a sequence of length \((L \times N_s + 1) \times C\). The sequence is then passed through a forward state space update. Owing to the aggregation behavior of SSMs, only the final token, the appended one, is retained as the output feature \(\hat{q}_j^m\).

\begin{table*}
\centering
\setlength{\tabcolsep}{4pt}
\begin{tabular}{cccccccccccccc}
\specialrule{1pt}{0pt}{0pt}
\multirow{2}{*}{Type}                                           & \multirow{2}{*}{Method} & \multirow{2}{*}{Venue} & \multirow{2}{*}{Modality} & \multicolumn{4}{c}{mAP(\%)}                              &  & \multicolumn{5}{c}{mAR(\%)}                                              \\ \cline{5-8} \cline{10-14} 
                                                                         &                                  &                                 &                                    & 0.5           & 0.75          & 0.95          & Avg.          &  & 100           & 50            & 20            & 10            & Avg.          \\ 
\specialrule{1pt}{0pt}{0pt}\addlinespace[2pt]
\multirow{2}{*}{\begin{tabular}[c]{@{}c@{}}Anchor-\\ based\end{tabular}} & BMN                              & ICCV'19                         & $\mathcal{V}$                      & 24.0          & 7.60           & 0.10           & 10.6          &  & 53.3          & 41.2          & 31.6          & 26.9          & 38.3          \\
                                                                         & BMN(I3D)                         & ICCV'19                         & $\mathcal{V}$                      & 10.6          & 1.70           & 0.00           & 4.10           &  & 48.5          & 44.4          & 37.1          & 31.6          & 40.4          \\ 
\hline\addlinespace[2pt]
\multirow{10}{*}{\begin{tabular}[c]{@{}c@{}}Anchor-\\ free\end{tabular}}  & AGT                              & Arxiv'21                        & $\mathcal{V}$                      & 17.9          & 9.40           & 0.10           & 9.10           &  & 43.2          & 34.2          & 24.6          & 16.7          & 29.7          \\
                                                                         & ActionFormer                     & ECCV'22                         & $\mathcal{V}$                      & 95.3          & 90.2          & 23.7          & 69.7          &  & 88.4          & 89.6          & 90.3          & 90.4          & 89.7          \\
                                                                         & TriDet                           & CVPR'23                         & $\mathcal{V}$                      & 96.3          & 86.8          & 23.6          & 68.9          &  & 91.0          & 90.4          & 90.0          & 88.7          & 90.0          \\
                                                                         & MDS                              & MM'20                           & $\mathcal{AV}$                     & 12.8          & 1.60           & 0.00           & 4.80           &  & 37.9          & 36.7          & 34.4          & 32.2          & 35.3          \\
                                                                         & AVFusion                         & Arxiv'21                        & $\mathcal{AV}$                     & 65.4          & 23.9          & 0.10           & 29.8          &  & 63.0          & 59.3          & 54.8          & 52.1          & 57.3          \\
                                                                         & BA-TFD                           & DICTA'22                        & $\mathcal{AV}$                     & 76.9          & 38.5          & 0.30           & 38.6          &  & 66.9          & 64.1          & 60.8          & 58.4          & 62.5          \\
                                                                         & BA-TFD+                          & CVIU'23                         & $\mathcal{AV}$                     & 96.3          & 85.0          & 4.40           & 61.9          &  & 81.6          & 80.5          & 79.4          & 78.8          & 79.8          \\
                                                                         & UMMAFormer                       & MM'23                           & $\mathcal{AV}$                     & \textbf{98.8} & \textbf{95.5} & \underline{37.6} & \textbf{77.3} &  & 92.4          & 92.5          & 92.5          & \underline{92.1} & \underline{92.3} \\
                                                                         & ELF-MDC & TCSVT'24 & $\mathcal{AV}$ & 94.9 & 74.9 & 1.90 & 57.2 & & 76.1 & 74.2 & 72.2 & 71.2 & 73.4 \\
                                                                         & DiMoDif                          & Arxiv'24                        & $\mathcal{AV}$                     & 95.5          & 87.9          & 20.6          & 67.8          &  & \textbf{94.2}          & \textbf{93.7} & \underline{92.7} & 91.4          & 91.9          \\ 
\hline\addlinespace[2pt]
\multirow{4}{*}{\begin{tabular}[c]{@{}c@{}}Query-\\ based\end{tabular}}  & TadTR                            & TIP'22                          & $\mathcal{V}$                      & 80.2          & 61.1          & 5.20          & 48.8          &  & 72.9          & 72.5          & 70.56         & 69.2          & 71.2          \\
                                                                         & TE-TAD                           & CVPR'24                         & $\mathcal{V}$                      & 85.7          & 64.9          & 7.13          & 52.6          &  & 79.2          & 78.6          & 78.4          & 76.2          & 77.9          \\

& \cellcolor{gray!30}FullFormer (Ours)               & \cellcolor{gray!30}--                               & \cellcolor{gray!30}$\mathcal{AV}$                     & \cellcolor{gray!30}94.6          & \cellcolor{gray!30}85.7          & \cellcolor{gray!30}29.4          & \cellcolor{gray!30}69.9          & \cellcolor{gray!30} & \cellcolor{gray!30}88.4 & \cellcolor{gray!30}88.4          & \cellcolor{gray!30}86.9          & \cellcolor{gray!30}85.5          & \cellcolor{gray!30}87.3          \\ 
                                                                         
                                                                         & \cellcolor{gray!30}DeformTrace (Ours)               & \cellcolor{gray!30}--                               & \cellcolor{gray!30}$\mathcal{AV}$                     & \cellcolor{gray!30}\underline{97.1}          & \cellcolor{gray!30}\underline{90.7}          & \cellcolor{gray!30}\textbf{38.1}          & \cellcolor{gray!30}\underline{75.3}          & \cellcolor{gray!30} & \cellcolor{gray!30}\underline{93.3} & \cellcolor{gray!30}\underline{93.1}          & \cellcolor{gray!30}\textbf{92.8}          & \cellcolor{gray!30}\textbf{92.3}          & \cellcolor{gray!30}\textbf{92.9}          \\ 
\specialrule{1pt}{0pt}{0pt}
\end{tabular}
\caption{Temporal forgery localization results on LAV-DF \cite{cai2023glitch} benchmark. The modality denotes the input type: \(\mathcal{V}\) for visual, \(\mathcal{A}\) for audio and \(\mathcal{AV}\) for both. The best results are \textbf{bolded}, and second-best \underline{underlined}.}
\label{table1}
\end{table*}

\subsection{Losses and Optimization}

In Relay Token Mechanism, we introduce two auxiliary losses: enhance loss and cooperation loss. The former encourages each relay token to better aggregate information from its neighboring sequence segments, while the latter promotes effective collaboration among relay tokens.

\textbf{Enhance loss.} For the \(k\)-th relay token \(r_k\), we compute the average representation \(f_k^{avg}\) of its neighboring non-relay subsequences (excluding other relay tokens). The enhance loss is then defined as:
\begin{equation}
\mathcal{L}_{\text{enh}} = -\frac{1}{N_r} \sum_{k=1}^{N_r} \cos\left(r_k, f_k^{avg}\right),
\label{eq:3}
\end{equation}

\textbf{Cooperation loss.} To promote diversity and reduce redundancy among relay tokens, we introduce the \textit{Cooperation Loss}, which minimizes the mutual information (MI) between different relay tokens. This can be approximated by encouraging the similarity matrix \(G\) of relay tokens to approach a scaled identity matrix:
\begin{equation}
\begin{aligned}
\mathcal{L}_{\text{coop}} &= \min \sum_{i < j} I(r_i; r_j), \\
&\approx \left\lVert G - \gamma I \right\rVert^2 = \sum_{i \ne j} G_{ij}^2 + \sum_{k=1}^{N_r} (G_{kk} - \gamma)^2,
\label{eq:4}
\end{aligned}
\end{equation}
where $\gamma$ is a hyperparameter controlling the target self-similarity, we set it to 1.

\textbf{Task-aware loss.} Following DETR~\cite{carion2020end}, we adopt the standard Hungarian Matching loss \(\mathcal{L}_{\text{match}}\), which includes classification loss \(\mathcal{L}_{\text{ce}}\) and regression loss \(\mathcal{L}_{\text{reg}}\). It is defined as:
\begin{equation}
\mathcal{L}_{\text{match}}(\mathcal{A}^M, \mathcal{A}) = \sum_{i=1}^{N_f} \sum_{u \in \{\text{ce}, \text{reg}\}} \mathcal{L}_u\left(\mathcal{A}^M_{\pi(i)}, \mathcal{A}_i\right),
\label{eq:5}
\end{equation}
where \(\pi\) denotes the optimal permutation from Hungarian Matching, \(\mathcal{A}^M, \mathcal{A}\) denote the predicted and ground-truth segments, respectively. Besides, we apply a video-level cross-entropy loss between the predicted label \(\hat{y}\) and ground truth \(y\), denoted as \(\mathcal{L}_{\text{cls}}\).

Overall, the training loss is a weighted combination of four components:
\begin{equation}
\mathcal{L} = \mathcal{L}_{\text{match}} + \mathcal{L}_{\text{cls}} + \lambda_1 \cdot \mathcal{L}_{\text{enh}} + \lambda_2 \cdot \mathcal{L}_{\text{coop}}.
\label{eq:6}
\end{equation}
where \(\lambda_1\) and \(\lambda_2\) are hyperparameters controlling the contributions of the enhance and cooperation losses, respectively.

\section{Experiments}

\subsection{Experiment Setup} 

\textbf{Datasets and Evaluation Metrics.} We conduct experiments on two audio-visual deepfake datasets: LAV-DF~\cite{cai2023glitch} and AV-Deepfake1M~\cite{cai2024av}. LAV-DF contains 78K/31K/26K train/val/test samples. AV-Deepfake1M is a larger and more challenging benchmark with 746K/57K/343K samples and finer-grained forgeries, posing greater difficulty for TFL models. Following standard practice, we report mean Average Precision (mAP) at IoU thresholds \(\{0.5, 0.75, 0.9, 0.95\}\), and mean Average Recall (mAR) at different numbers of proposals \(\{5, 10, 20, 30, 50, 100\}\). Additionally, we evaluate video-level detection on AV-Deepfake1M using Area Under the Curve (AUC) as the metric. We compare model parameters and computational cost on a single RTX 3090 GPU, evaluating all methods on the full LAV-DF test set (avg. video length: 8.6 s) with total/trainable parameters, FLOPs, and inference time.

\textbf{Implementation Details.} We adopt Raven’s visual and audio encoders \cite{haliassos2023jointly} for feature extraction, due to their strong performance on diverse downstream tasks. Specifically, we use the base model pretrained on VoxCeleb2 \cite{chung2018voxceleb2} and LRS3 \cite{afouras2018lrs3}. Videos are sampled at 25 fps and audio at 16 kHz. Visual frames are resized to \(224 \times 224\) pixels, and audio is converted into 80-bin mel-spectrograms using a 40 ms Hamming window with a 40 ms hop size. Each input spans 8 seconds, resulting in \(T = N_1 = 200\) frames and \(C=256\) dimensions after encoding. We adopt \(L = 6\) feature levels via 1D convolution with kernel size 3, and use \(M = 3\) decoder layers. In each deformable self-/cross-SSM layer, we sample \(N = 6\) points per scale. We set the number of query and relay tokens to \(N_q = 60\) and \(N_r = 8\), respectively. The total loss combines two terms weighted by \(\lambda_1 = 0.5\) and \(\lambda_2 = 0.2\). We train for 30 epochs on AV-Deepfake1M and 100 epochs on LAV-DF with a batch size of 32, using AdamW \cite{loshchilov2017decoupled} and a cosine scheduler with 5-epoch warm-up. The learning rate is set to 2e-4. All experiments are conducted on eight NVIDIA RTX 3090 GPUs. 

\textbf{Comparison Methods.} For the temporal forgery localization task, our baseline systems include three publicly available audio-visual TFL methods: BA-TFD \cite{cai2022you}, BA-TFD+ \cite{cai2023glitch}, and UMMAFormer \cite{zhang2023ummaformer}. In addition, we construct a baseline by replacing all SSMs in DeformTrace with Transformer blocks, referred to as FullFormer, for comparison. To enable a comprehensive comparison, we also report results on the LAV-DF and AV-Deepfake1M benchmarks. 

\begin{table*}
\centering
\small
\setlength{\tabcolsep}{3.5pt}
\begin{tabular}{cccccccccccccccc}
\specialrule{1pt}{0pt}{0pt}
\multirow{2}{*}{Type}                                                    & \multirow{2}{*}{Method}  & \multirow{2}{*}{Venue} & \multirow{2}{*}{Modality} & \multicolumn{5}{c}{mAP(\%)}          &                      & \multicolumn{6}{c}{mAR(\%)}                 \\ \cline{5-9} \cline{11-16} 
                                                                         &                          &                        &                           & 0.5  & 0.75 & 0.9  & 0.95 & Avg. &                      & 50   & 30   & 20   & 10   & 5    & Avg. \\ 
\specialrule{1pt}{0pt}{0pt}\addlinespace[2pt]
\multirow{11}{*}{\begin{tabular}[c]{@{}c@{}}Anchor-\\ free\end{tabular}} & Meso4                    & WIFS'18                & $\mathcal{V}$             & 9.86 & 6.05 & 2.22 & 0.59 & 4.68 &                      & 38.9 & 38.9 & 38.8 & 36.5 & 26.9 & 36.0 \\
                                                                         & MesoInception4           & WIFS'18                & $\mathcal{V}$             & 8.50 & 5.16 & 1.89 & 0.50 & 4.01 &                      & 39.3 & 39.2 & 39.0 & 35.8 & 24.6 & 35.6 \\
                                                                         & EfficientViT             & ICIAP'22               & $\mathcal{V}$             & 14.7 & 2.42 & 0.13 & 0.01 & 4.32 &                      & 27.1 & 27.0 & 26.4 & 23.9 & 20.3 & 24.9 \\
                                                                         & TriDet+VideoMAEv2        & CVPR'23                & $\mathcal{V}$             & 21.7 & 5.83 & 0.54 & 0.06 & 7.03 &                      & 20.3 & 20.2 & 20.1 & 19.5 & 18.2 & 19.7 \\
                                                                         & TriDet+InternVideo       & CVPR'23                & $\mathcal{V}$             & 29.7 & 9.02 & 0.79 & 0.09 & 9.89 &                      & 24.1 & 24.1 & 24.0 & 23.5 & 22.6 & 23.6 \\
                                                                         & ActionFormer+VideoMAEv2  & ECCV'22                & $\mathcal{V}$             & 20.2 & 5.73 & 0.57 & 0.07 & 6.65 &                      & 20.0 & 19.9 & 19.8 & 19.1 & 17.8 & 19.3 \\
                                                                         & ActionFormer+InternVideo & ECCV'22                & $\mathcal{V}$             & 36.1 & 12.0 & 1.23 & 0.16 & 12.4 &                      & 27.1 & 27.1 & 27.0 & 26.6 & 25.8 & 26.7 \\
                                                                         & BA-TFD                   & DICTA'22               & $\mathcal{AV}$            & 37.4 & 06.3 & 0.19 & 0.02 & 11.0 &                      & 45.6 & 40.4 & 36.0 & 30.7 & 26.8 & 35.9 \\
                                                                         & BA-TFD+                  & CVIU'23                & $\mathcal{AV}$            & 44.4 & 13.6 & 0.48 & 0.03 & 14.6 &                      & 48.9 & 44.5 & 40.4 & 34.7 & 29.9 & 39.7 \\
                                                                         & UMMAFormer               & MM'23                  & $\mathcal{AV}$            & 51.6 & 28.1 & 7.65 & 1.58 & 22.2 &                      & 44.1 & 43.9 & 43.5 & 42.1 & 40.3 & 42.8 \\
                                                                         & DiMoDif                  & Arxiv'24               & $\mathcal{AV}$            & \underline{86.9} & \underline{76.0} & \underline{28.7} & \underline{5.43} & \underline{49.3} &                      & \underline{81.6} & \underline{80.9} & \underline{80.3} & \underline{78.8} & \underline{76.6} & \underline{79.6} \\ 
\hline\addlinespace[2pt]
\multirow{3}{*}{\begin{tabular}[c]{@{}c@{}}Query-\\ based\end{tabular}}  & TadTR                    & TIP'22                 & $\mathcal{V}$             & 60.5 & 33.8 & 5.12 & 1.20 & 25.2 & \multicolumn{1}{c}{} & 57.6 & 55.8 & 54.2 & 51.2 & 50.0 & 53.8 \\
                                                                         & TE-TAD                   & CVPR'24                & $\mathcal{V}$             & 71.4 & 54.4 & 11.3 & 2.31 & 34.9 & \multicolumn{1}{c}{} & 66.8 & 64.7 & 64.0 & 61.4 & 59.9 & 63.3 \\

& \cellcolor{gray!30}FullFormer (Ours)        & \cellcolor{gray!30}--                     & \cellcolor{gray!30}$\mathcal{AV}$            & \cellcolor{gray!30}77.3 & \cellcolor{gray!30}57.9 & \cellcolor{gray!30}22.2 & \cellcolor{gray!30}4.12 & \cellcolor{gray!30}40.4 & \cellcolor{gray!30} & \cellcolor{gray!30}70.8 & \cellcolor{gray!30}67.5 & \cellcolor{gray!30}66.8 & \cellcolor{gray!30}65.8 & \cellcolor{gray!30}63.0 & \cellcolor{gray!30}66.8 \\ 
                                                                         
                                                                         & \cellcolor{gray!30}DeformTrace (Ours)        & \cellcolor{gray!30}--                     & \cellcolor{gray!30}$\mathcal{AV}$            & \cellcolor{gray!30}\textbf{92.0} & \cellcolor{gray!30}\textbf{78.7} & \cellcolor{gray!30}\textbf{31.2} & \cellcolor{gray!30}\textbf{9.58} & \cellcolor{gray!30}\textbf{52.9} & \cellcolor{gray!30} & \cellcolor{gray!30}\textbf{86.2} & \cellcolor{gray!30}\textbf{84.2} & \cellcolor{gray!30}\textbf{81.4} & \cellcolor{gray!30}\textbf{79.0} & \cellcolor{gray!30}\textbf{78.0} & \cellcolor{gray!30}\textbf{81.8} \\ 
\specialrule{1pt}{0pt}{0pt}
\end{tabular}
\caption{Temporal forgery localization results on AV-Deepfake1M \cite{cai2024av} benchmark. The modality denotes the input type: \(\mathcal{V}\) for visual, \(\mathcal{A}\) for audio and \(\mathcal{AV}\) for both. The best results are \textbf{bolded}, and second-best \underline{underlined}.}
\label{table2}
\end{table*}

\subsection{Comparison with State-of-the-arts}

\textbf{TFL results.} Tables~\ref{table1} and~\ref{table2} compare our method with state-of-the-art TFL approaches on LAV-DF and AV-Deepfake1M. On both datasets, our method outperforms the pure Transformer baseline by over 7\% in mAP and mAR, highlighting the effectiveness of SSMs and our proposed modules. On LAV-DF, it achieves top results on mAP@0.95 and mAP@\{20, 10, Avg.\}, and ranks second on the remaining metrics, demonstrating robustness across evaluation criteria. On AV-Deepfake1M, leading models like UMMAFormer and BA-TFD+ suffer performance drops due to greater data diversity and shorter segments. In contrast, our method consistently delivers state-of-the-art results, outperforming all existing approaches across all metrics with significant margins. Specifically, DeformTrace surpasses the second-best model, DiMoDif, by 5.1\%, 2.7\%, 2.5\%, and 4.15\% in mAP@\{0.5, 0.75, 0.9, 0.95\}, respectively—yielding an average improvement of 3.6\%. In terms of mAR under 50, 30, 20, 10, and 5 proposals, DeformTrace achieves gains of 4.6\%, 3.3\%, 1.1\%, and 1.4\%, averaging a 2.2\% improvement over DiMoDif. These results clearly demonstrate the model’s strength in precisely localizing fine-grained forgery segments, even under challenging conditions. 

\begin{table}[t]
\centering
\small
\setlength{\tabcolsep}{2.7pt}
\begin{tabular}{ccccc}
\specialrule{1pt}{0pt}{0pt}\addlinespace[2pt]
Method             & Total{[}M{]} & Train{[}M{]} & FLOPs{[}G{]} & Time{[}ms{]} \\ \specialrule{1pt}{0pt}{0pt}\addlinespace[2pt]
BA-TFD             & \textbf{5.5}                  & \textbf{5.5}                  & 948.1                & \underline{605}                  \\
BA-TFD+            & 152.9                & 152.9                & \underline{218.2}                & 681                  \\
UMMAFormer         & 165.9                & 49.72                & 1563.9               & 857                  \\

\cellcolor{gray!30}FullFormer (Ours) & \cellcolor{gray!30}\underline{116.7}                & \cellcolor{gray!30}\underline{19.2}                 & \cellcolor{gray!30}233.7                & \cellcolor{gray!30}\underline{126}                  \\ 

\cellcolor{gray!30}DeformTrace (Ours) & \cellcolor{gray!30}118.3                & \cellcolor{gray!30}20.8                 & \cellcolor{gray!30}\textbf{212.4}                & \cellcolor{gray!30}\textbf{104}                  \\ 
\specialrule{1pt}{0pt}{0pt}
\end{tabular}
\caption{Model size and computation comparison of \(\mathcal{AV}\) TFL methods. \textit{Total}, \textit{Train}, and \textit{Time} denote total parameters, trainable parameters, and per-video inference time.}
\label{table3}
\end{table}

\textbf{Efficiency results.} We compare the model size and computational cost of audio-visual TFL methods, reporting both total and trainable parameters to consider differences in feature extractors. Computation cost and inference time includes feature extraction. As shown in Table~\ref{table3}, our model is more efficient and lightweight: it reduces trainable parameters by 28.92M and 132.1M compared to UMMAFormer and BA-TFD+, and achieves the lowest GFLOPs—6.4\(\times\) less than UMMAFormer. It also runs 7.3\(\times\) faster than UMMAFormer and 5.8\(\times\) faster than BA-TFD and BA-TFD+. Compared to our Baseline, DeformTrace slightly increases parameter count but significantly reduces computation and inference time. This efficiency benefits from the query-based architecture (fully end-to-end without post-processing) and the linear complexity of SSMs. The model’s computational cost is mainly concentrated in the sequence processing modules, where the linear-complexity SSMs account for approximately a 4:1 cost ratio between DS-SSM and DC-SSM, while the lightweight MLP-based offset prediction has a negligible contribution. Overall, DeformTrace offers a better trade-off between performance and efficiency. 


\begin{figure*}[t]
\centering
\includegraphics[width=0.98\textwidth]{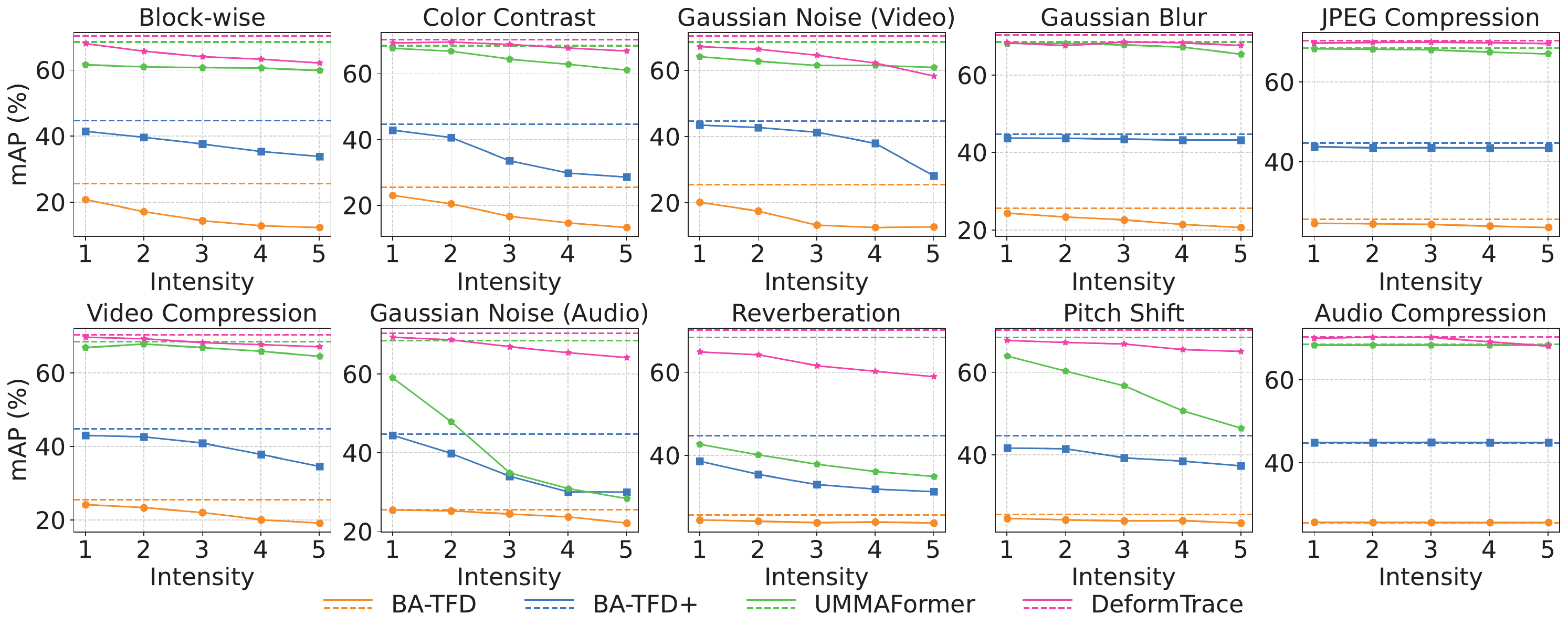}
\caption{Robustness evaluation under various compression and degradation scenarios. The experiments include 6 visual distortions (Block-wise, Color Contrast, Gaussian Noise (video), Gaussian Blur, JPEG Compression and Video Compression) and 4 audio distortions (Gaussian Noise (audio), Reverberation, Pitch Shift and Audio Compression). In the figure, colors denote methods; solid lines show mAP across intensities, dashed lines show mAP on clean videos. DeformTrace achieves the highest mAP on clean videos and demonstrates strong robustness across various distortion scenarios.}
\label{robustness}
\end{figure*}

\begin{table}[ht]
\centering
\small
\setlength{\tabcolsep}{4.5pt}
\begin{tabular}{cccccccc}
\specialrule{1pt}{0pt}{0pt}
\#  &DS-SSM                 & \multicolumn{1}{c|}{DC-SSM}                 & \(\mathcal{L}_{\text{enh}}\) & \multicolumn{1}{c|}{\(\mathcal{L}_{\text{coop}}\)} & mAP  & mAR  & AUC  \\
\specialrule{1pt}{0pt}{0pt}\addlinespace[2pt]
\multicolumn{8}{c}{\textit{Vanilla SSM}}                                                                                                                      \\ 
\hline
1  &\xmark & \multicolumn{1}{c|}{\xmark} & \xmark        & \multicolumn{1}{c|}{\xmark}         & 41.2 & 68.7 & 90.3 \\ \hline\addlinespace[2pt]
\multicolumn{8}{c}{\textit{Deformable SSM}}                                                                                                                \\ 
\hline
2  &\cmark & \multicolumn{1}{c|}{\xmark} & \xmark        & \multicolumn{1}{c|}{\xmark}         & 44.4 & 73.2 & 95.7 \\
3  &\xmark & \multicolumn{1}{c|}{\cmark} & \xmark        & \multicolumn{1}{c|}{\xmark}         & 47.8 & 75.7 & 97.3 \\

4  &\cmark & \multicolumn{1}{c|}{\cmark} & \xmark        & \multicolumn{1}{c|}{\xmark}         & 49.8 & 78.2 & 98.1 \\ \hline\addlinespace[2pt]
\multicolumn{8}{c}{\textit{Relay Token Mechanism}}                                                                                                         \\ 
\hline
5  &\cmark & \multicolumn{1}{c|}{\cmark} & \cmark        & \multicolumn{1}{c|}{\xmark}         & \underline{51.4} & \underline{79.2} & 98.3 \\
6  &\cmark & \multicolumn{1}{c|}{\cmark} & \xmark        & \multicolumn{1}{c|}{\cmark}         & 51.7 & 78.9 & \underline{98.5} \\
\cellcolor{gray!30}7  &\cellcolor{gray!30}\cmark & \multicolumn{1}{c|}{\cellcolor{gray!30}\cmark} & \cellcolor{gray!30}\cmark        & \multicolumn{1}{c|}{\cellcolor{gray!30}\cmark}         & \cellcolor{gray!30}\textbf{52.9} & \cellcolor{gray!30}\textbf{81.8} & \cellcolor{gray!30}\textbf{99.2} \\ 
\specialrule{1pt}{0pt}{0pt}
\end{tabular}
\caption{Ablation study results on AV-Deepfake1M.}
\label{table5}
\end{table}

\subsection{Ablation Study}

\textbf{Ablation studies on DeformTrace.} To assess each component's impact, we conduct ablations on the AV-Deepfake1M full test set, as shown in Table~\ref{table5}. Compared to the baseline (Line 1) that replaces all Deformable SSMs with vanilla SSMs and removes relay losses, DeformTrace (Line 7) improves mAP by 28.4\%, mAR by 19.1\%, and AUC by 9.9\%, highlighting the effectiveness of all modules. Specifically, Lines 2 and 3 show that both DS-SSM and DC-SSM enhance performance, with DC-SSM providing larger gains due to: 1) without relay tokens, DS-SSM suffers more from long-range decay, and 2) DC-SSM enhances query–feature sequence interaction over vanilla SSM, which is crucial for proposal refinement. Line 4 shows that combining both modules boosts performance further. Comparing Line 4 with Lines 5–6 indicates that each relay loss brings moderate gains, while using both (Line 7) yields the best results, surpassing Line 4 by 3.1\% mAP, 3.6\% mAR, and 1.1\% AUC.

\begin{figure}[t]
\centering
\includegraphics[width=\columnwidth]{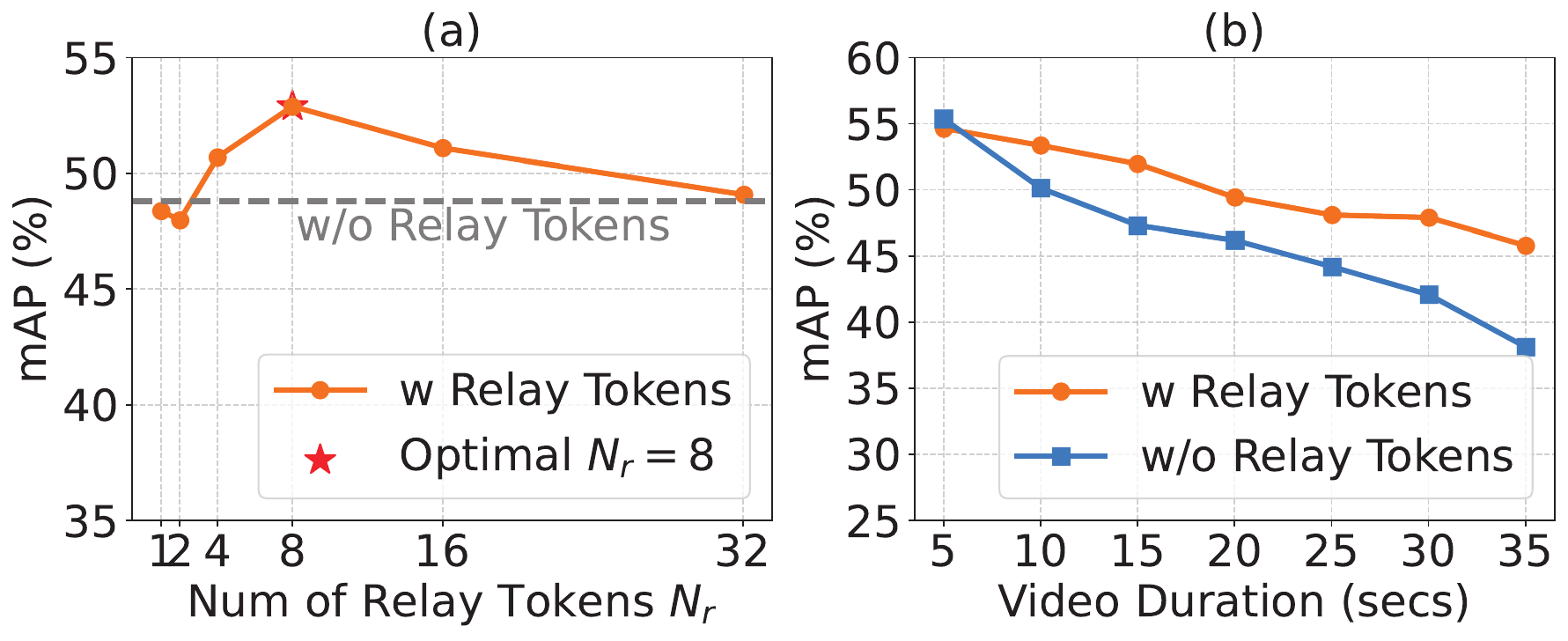}
\caption{(a) Ablation study on the number of relay tokens (avg. video length: 9s). (b) Performance vs. video duration.}
\label{ablation}
\end{figure}

\textbf{Ablation studies on hyperparameters.} We conduct an ablation study on the number of relay tokens \(N_r\), using the AV-Deepfake1M full test set with an average video duration of 9 seconds. As shown in Figure \ref{ablation}(a), when \(N_r = 1\) or \(2\), performance is comparable to or worse than the baseline without relay tokens, as each sub-state space still contains too many tokens to alleviate long-range decay. As \(N_r\) increases, mAP improves significantly, peaking at \(N_r = 8\), where token distribution per subspace aligns well with decoder queries \(N_q\). Beyond this, further increasing \(N_r\) leads to performance drop due to over-segmentation, which impedes global information flow and raises complexity.

\textbf{Performance vs. Video Duration.} 
We select seven specific durations ranging from 5 to 35 seconds and construct sub-datasets by sampling 1,000 videos closest to each target duration from AV-Deepfake1M. This setup allows us to systematically examine how model performance varies with sequence length. As shown in Figure~\ref{ablation}(b), model with relay tokens outperform its counterparts as duration increases, demonstrating a stronger capacity to capture long-range dependencies. In contrast, model without relay tokens shows marginally better performance on short videos, but their performance deteriorates rapidly on longer sequences. The results highlight the importance of relay tokens in maintaining robust temporal reasoning over time.

\subsection{Robustness Study}

To evaluate the robustness of our model, we sampled 2,500 videos (2000 from LAV-DF, 500 from AV-Deepfake1M) and applied 10 common transmission perturbations (6 visual, 4 audio) at 5 intensity levels each, yielding 125,000 degraded videos. We tested our model alongside three audio-visual TFL methods, comparing performance using mAP. As shown in Figure~\ref{robustness}, DeformTrace achieves the highest mAP on clean videos and exhibits strong robustness against Color Contrast, Gaussian Blur, Pitch Shift and three compression types. While performance decreases at high intensities for other distortions, our approach consistently outperforms baselines, demonstrating superior robustness.

\section{Conclusion and Discussion}

We propose DeformTrace, a novel TFL framework that integrates State Space Models (SSMs) for a better performance-efficiency trade-off. It includes a Deformable Self-SSM (DS-SSM) with dynamic receptive fields for precise boundary localization, a Deformable Cross-SSM (DC-SSM) for modeling cross-sequence interactions in sparse forgeries, and a Relay Token Mechanism to mitigate long-range information decay. By combining the global modeling of Transformers with the efficiency of SSMs, DeformTrace achieves precise localization with significantly faster inference, exceeding previous methods in both precision and robustness.

While DC-SSM is used in our work solely for modeling between forgery queries and features, it can generalize to any two independent sequences such as audio and video for audio-visual correspondence learning, enabling broader applications beyond temporal forgery localization.

\section{Acknowledgments}
This research is funded in part by the National Natural Science Foundation of China (62171326, 62371350, 62471343), Key Science and Technology Research Project of Xinjiang Production and Construction Corps (2025AB029) in 2025, Guangdong OPPO Mobile Telecommunications Corp. and Wuhan University Supercomputing Center.



\bibliography{aaai2026}

\end{document}